# In-Pixel Foreground and Contrast Enhancement Circuits with Customizable Mapping

Md Rahatul Islam Udoy[1], Md Mazharul Islam[1], Elijah Johnson[2], and Ahmedullah Aziz[1*]

[1]Department of Electrical Engineering and Computer Science, University of Tennessee, Knoxville, TN 37996, USA
[2]Department of Electrical Engineering, Columbia University, New York, New York, 10027, USA
*Corresponding Author's Email: aziz@utk.edu

**This paper presents an innovative in-pixel contrast enhancement circuit that performs image processing directly within the pixel circuit. The circuit can be tuned for different modes of operation. In foreground enhancement mode, it suppresses low-intensity background pixels to nearly zero, isolating the foreground for better object visibility. In contrast enhancement mode, it improves overall image contrast. The contrast enhancement function is customizable both during the design phase and in real-time, allowing the circuit to adapt to specific applications and varying lighting conditions. A model of the designed pixel circuit is developed and applied to a full pixel array, demonstrating significant improvements in image quality. Simulations performed in HSPICE show a nearly 6x increase in Michelson Contrast Ratio (CR) in the foreground enhancement mode. The simulation results indicate its potential for real-time, adaptive contrast enhancement across various imaging environments.**

*Index Terms— in-pixel processing, image sensor, pixel-level processing, contrast enhancement, HyperFET, PTM, background suppression, contrast stretching.*

In digital imaging, enhancing image quality is crucial, particularly in applications where the precise differentiation of foreground objects from the background is essential [1]. Contrast is the difference in luminance that distinguishes objects plays a key role in achieving this clarity [2]. However, many real-world scenarios, such as low-light or high-noise environments, produce low-contrast images where the intensity differences between important features and their backgrounds are minimal[3]. This hinders the ability of computer vision systems to detect critical details, posing challenges in fields like surveillance, medical diagnostics, and autonomous systems[4]. Foreground enhancement techniques address this issue by isolating key objects within the image, reducing the prominence of background elements and improving object visibility [5]. This selective emphasis enhances decision-making and improves the accuracy of machine learning tasks such as object detection and image segmentation [6].

Traditionally, contrast and foreground enhancement are performed in external processing units after the image is captured[7]. In conventional systems, these external processors cannot handle all the pixels in a pixel array simultaneously, creating bottlenecks that slow down image processing[8]. This sequential handling of pixel data reduces the system's overall speed and limits real-time performance[9]. In contrast, in-pixel processing allows all pixels to be processed in parallel, eliminating bottlenecks and significantly improving processing speed and efficiency[8]. In-pixel processing also enhances security. External processing units are often prime targets for cyberattacks, where raw image data can be intercepted, leaked, or tampered with [10]. By embedding critical image processing functions directly within the sensor, the attack surface is significantly reduced. Fewer external components are involved in handling raw data, minimizing the risk of data breaches and ensuring that sensitive information stays within the sensor for as long as possible. Moreover, in traditional systems, the raw signal from the sensor must travel a considerable distance to reach the external processor[9]. During this transmission, the signal is vulnerable to noise, which degrades image quality even before any enhancement occurs. In-pixel processing solves this problem by reducing the distance raw data must travel, thus minimizing noise interference. While in-pixel processing offers significant advantages, it is not a straightforward task of transferring processing circuitry from the processing unit into the pixel. Such an approach would result in increased pixel size, thereby reducing the achievable resolution of the pixel array. Therefore, a meticulous design of the pixel circuitry is essential to integrate processing capabilities while maintaining compactness. Several studies in literature have explored in-pixel processing techniques, reflecting growing interest in enhancing imaging performance directly at the sensor level [8,11,12].

This paper presents a novel in-pixel contrast enhancement circuit that addresses the aforementioned limitations by performing real-time image enhancement directly within the pixel circuit. We create a model of the pixel circuit, enabling us to simulate the effect on all pixels of an array. The circuit operates in two modes: in foreground enhancement mode, it suppresses low-intensity background pixels to highlight objects in the foreground, while in contrast enhancement mode, it optimizes object differentiation by adjusting intensity levels across the image. By embedding these functions within the pixel structure, the proposed design allows for efficient, parallel processing that minimizes bottlenecks, reduces the attack surface for cyber threats, and minimizes noise interference. Simulations demonstrate a significant improvement in image quality, including a marked increase in the Michelson Contrast Ratio (CR) under foreground enhancement conditions. This work offers a secure, adaptable, and efficient solution for modern imaging systems where real-time performance and image clarity are critical. In this paper, we present our discussion in the following way- the first section presents a fundamental understanding of contrast enhancement, pixel circuit, and a previously reported device with unique properties called HyperFET, which we use in our circuit design. The next section describes our designed circuit structure, its working principle and the approach of modeling the pixel circuit. The fourth section explains the customization technique of the transfer curves. Finally, in the last section, we present a variation analysis.

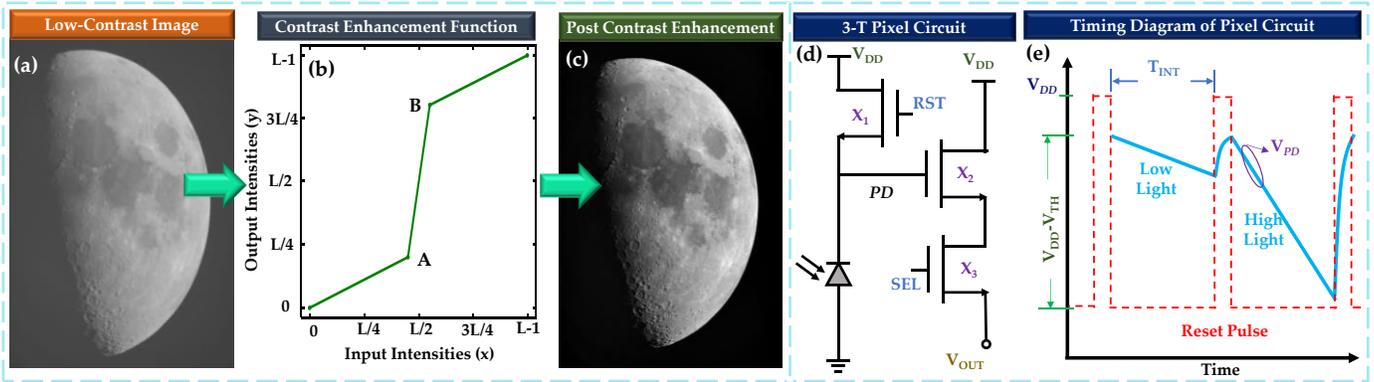

**Figure 1.** Contrast enhancement and pixel circuit. (**a**) A low-contrast image before applying the contrast enhancement function. (**b**) A contrast enhancement function, where points A and B are used to shape the transfer function. (**c**) Image after applying contrast enhancement. Image features are more visible than the low-contrast image. (**d**) Schematic of a basic 3-transistor (3-T) pixel circuit. Here, $X_1$, $X_2$, and $X_3$ are reset, source-follower, and selector transistors, respectively. (**e**) Timing diagram of a 3-T pixel circuit. Here, the slope of $V_{PD}$ is higher in the case of higher illumination.

**Background: Contrast enhancement, pixel circuit, and HyperFET**

Contrast is the difference in luminance or color that makes an object distinguishable from others in an image. In low-contrast images, the intensity levels between objects and the background are very similar, making it difficult to see details[13,14]. Contrast enhancement is a simple image enhancement technique that improves the contrast of an image by expanding the range of intensity levels [15]. In the original low-contrast image (Fig. 1(a)), details are not clearly visible due to limited intensity differences. The contrast enhancement function (Fig. 1 (b)) transforms the pixel intensities into new values[16]. As a result, the post-contrast enhancement image (Fig. 1(c)) shows enhanced contrast, making features more visible and improving overall image clarity. The contrast enhancement function shown in Fig. 1(b) is a piecewise linear transformation function. The shape of this function is customized by moving points A and B in both the x and y directions of the graph. Based on this transformation function, the mapping between input and output intensity changes. This type of transformation is typically done for low contrast images in an external processing unit, not inside the pixel array of an image sensor.

The main part of an image sensor chip is a 2D pixel array[17]. The basic building unit of a pixel array is typically a 3-transistor (3-T) pixel circuit as shown in Fig. 1(d)[18]. This circuit consists of a reverse-biased photodetector and three transistors: $X_1$, $X_2$, and $X_3$, which are referred to as the reset, source follower, and pixel selector transistors, respectively. The timing diagram for this circuit is presented in Fig. 1(e). The circuit operation begins with turning on the reset transistor $X_1$, which resets the photodetector (*PD*) node voltage ($V_{PD}$) to $V_{DD} - V_{TH,X1}$, in the case of a soft reset, where $V_{TH,X1}$ is the threshold voltage of the reset transistor. With a hard reset or by using a PMOS as the reset transistor, the PD node can be fully charged to $V_{DD}$ [19]. Once the reset phase ends, the integration period begins when the reset transistor is turned off. The voltage at the PD node decreases over time due to the photocurrent generated by photoelectrons in the photodetector, as described by the following equation:

$$\frac{dv}{dt} = \frac{I_{PD}}{C_{PD}} \quad (1)$$

where, $I_{PD}$ is the photocurrent and $C_{PD}$ is the intrinsic capacitance of the photodiode. Higher illumination leads to a higher $I_{PD}$, causing faster drop of $V_{PD}$, which is depicted in the second cycle of Fig. 1(e). $X_2$, which functions as a source follower transistor, is used to transfer the PD node signal while maintaining the accumulated charge undisturbed. The third transistor ($X_3$) functions as the pixel selector, allowing the pixel to be read from the array. $V_{OUT}$ is the output signal of this single 3-T pixel circuit.

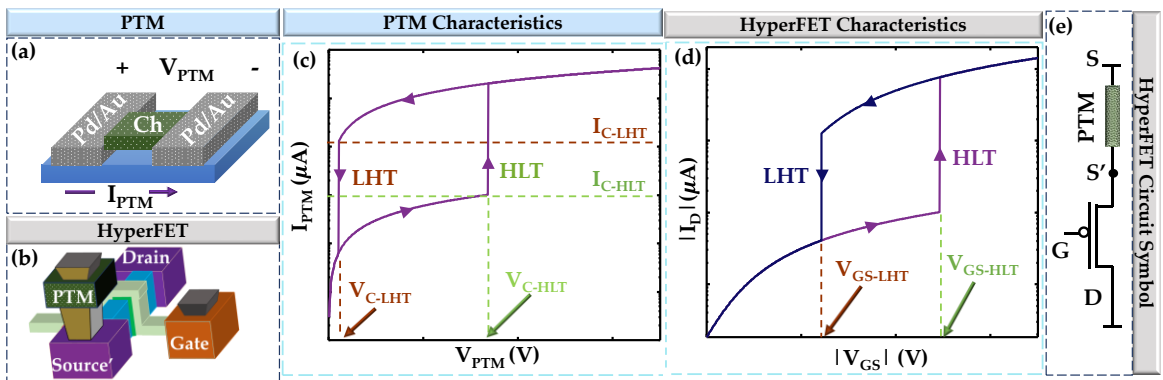

**Figure 2.** Introduction to PTM and HyperFET. (**a**) PTM structure. Here, Pd/Au is the metal electrode, and the green zone is the channel. (**b**) HyperFET structure, which can be achieved by integrating PTM in the source terminal of a MOSFET (**c**) Current vs. voltage characteristics of a PTM. (**d**) $|I_D|$ vs $|V_{GS}|$ of the HyperFET. Here, $I_D$ is the drain current, and $V_{GS}$ is the gate to source voltage of the HyperFET. (**e**) A P-type HyperFET circuit symbol.

In our pixel circuit design, we use Phase Transition Material (PTM) device, which has unique properties. PTMs are a class of materials known for their sharp resistivity changes, as documented in studies[20,21]. These transitions can be triggered by a variety of stimuli, including electrical [22,23], optical[24], or mechanical [25,26]. In this article, we use PTMs activated by electrical stimulus. The structure of PTM is illustrated in Fig. 2(a). At lower voltages, PTMs remain in a high resistance state (HRS). However, when the applied voltage exceeds a critical value ($V_{C-HLT}$), a high-to-low transition (HLT) of resistance occurs, as shown in Fig. 2(c)[27] and the PTM goes to low resistance state (LRS). The corresponding current at this transition point is referred to as $I_{C-HLT}$. On the other hand, when the voltage decreases below a certain threshold ($V_{C-LHT}$), the material goes through a low-to-high transition (LHT) of resistance [20], and the current at this point is called $I_{C-LHT}$. Various PTMs have been discovered, showing a wide range of transition voltages [28]. By integrating a PTM into the source terminal of a conventional Field-Effect Transistor (FET), as shown in Fig. 2(b), an interesting device known as the HyperFET is created [29]. This device combines the characteristics of a standard FET with the abrupt switching behavior of PTMs. At low $|V_{GS}|$, the FET remains off, and the PTM is in HRS, acting as a high resistance at the source terminal (Fig. 2(d) & (e)). When $|V_{GS}|$ exceeds a critical value ($|V_{GS-HLT}|$), the PTM undergoes an HLT, significantly lowering its resistance at the source terminal. Conversely, when $|V_{GS}|$ decreases below a critical value ($|V_{GS-LHT}|$), the PTM goes through LHT, returning to the HRS. Several useful circuits leveraging HyperFET technology have been demonstrated [11,30,31].

## IPFE circuit & pixel model

We design the In-Pixel Foreground Enhancement (IPFE) circuit to enhance contrast at the pixel level by applying a threshold-based filtering mechanism. The circuit configuration is shown in Fig. 3(a). We use a PMOS as the reset transistor ($X_1$) so that we can reset the PD node voltage to full $V_{DD}$. The PTM and the $X_2$ PMOS construct a HyperFET, which is used to collect the signal from the PD node. $X_3$ is the typical pixel selector transistor. The load transistor resides outside the pixel circuit. We simulate this circuit in HSPICE (an industry-grade simulator) by Synopsis[32]. To simulate the transistors, we use the NMOS/PMOS models of the IBM 65 nm 10LPe process and $V_{DD}$ is set to 1.2 V. The PTM we use is Pt/NbO$_2$/Pt [33,34] and to simulate the device, we calibrate a SPICE-based compact model reported in [20]. Simulation parameters for the PTM are presented in Table 1.

**Table 1: Simulation Parameters of the PTM.**

| Parameter | Definition | Value |
|---|---|---|
| $L_{PTM}$ | Length towards current flow | 5 nm |
| $A_{PTM}$ | Cross-sectional area | 27.5×27.5 nm$^2$ |
| $I_{C-HLT}$ | Critical current for HLT | 7.4 $\mu$A |
| $I_{C-LHT}$ | Critical current for LHT | 100 $\mu$A |
| $R_{HRS}$ | Resistance in HRS | 120.5 k$\Omega$ |
| $R_{LRS}$ | Resistance in LRS | 6.5 k$\Omega$ |

The circuit operation starts with turning on $X_1$, which resets the PD node to $V_{DD}$, which makes the HyperFET $V_{GS-HYP}=0$ and the PTM remains in the HRS. As $V_{DD}$ is shared among the PTM, $X_2$, $X_3$, and the load transistor, according to the voltage divider rule, voltage across the load transistor ($V_{OUT}$) will be very low due to HRS of the PTM. After the reset phase, $X_1$ is turned off. At the light integration phase, the voltage at the PD node starts to drop due to illumination in the photodiode according to equation (1) and $V_{GS\text{-}HYP}$ starts to increase. If $V_{GS\text{-}HYP}$ crosses $V_{GS\text{-}HLT}$, the PTM goes through HLT and shifts to LRS (see Fig. 2(d)). According to the voltage division rule, the voltage across the load transistor ($V_{OUT}$) becomes high. After the integration phase, the PD node voltage is read out by activating the selector transistor. Then the reset phase of the second cycle starts which again resets the PD node where $V_{GS\text{-}HYP}$ goes back to 0 V and reverts the PTM to HRS. We calculate the power delay product of the circuit = 2.89 fJ, which includes the energy during the reset phase, integration phase and readout phase. The I-V characteristics shape of the HyperFET is directly reflected in the $V_{OUT}$ signal.

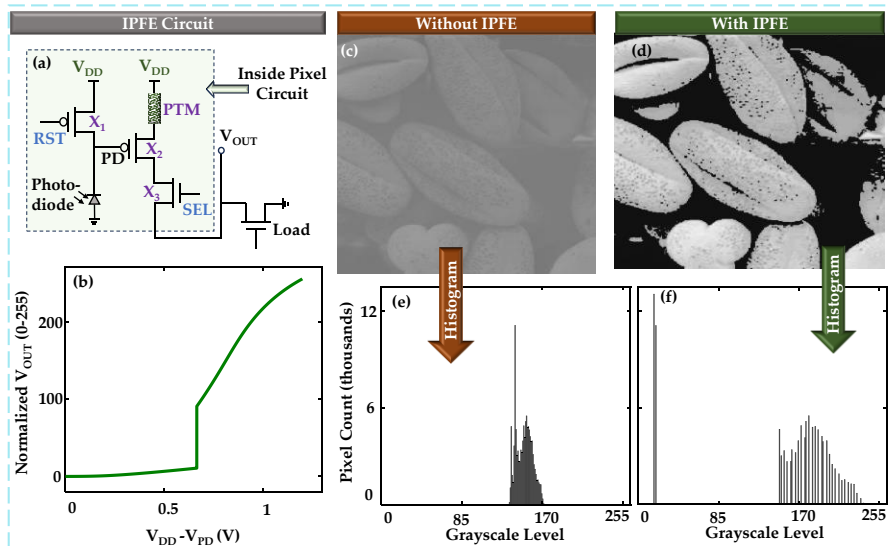

**Figure 3.** In-pixel foreground enhancement (IPFE) circuit. **(a)** Schematic of the proposed circuit. Here, $X_2$ and PTM form a p-type HyperFET. **(b)** Normalized output of the circuit between 0-255 levels (for 8-bit encoding) vs. the voltage drop at the PD node. **(c)** Low-contrast image without IPFE, where many features are not visible **(d)** Effect of IPFE on the image; here, features are visible. **(e)** Histogram of the low-contrast image, which shows a narrow distribution of grayscale levels. **(f)** Histogram of the image with IPFE. Here, the distribution is much broader.

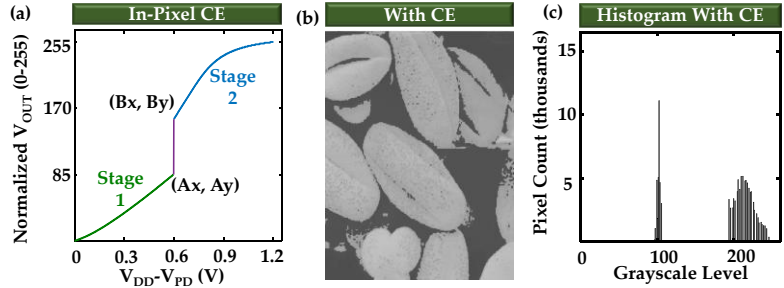

**Figure 4.** Contrast enhancement mode. **(a)** The input vs output curve of the pixel model for contrast enhancement (CE) mode. Stage 1 and Stage 2 are due to HRS and LRS, respectively. Ax, Ay, Bx, and By are tuned to achieve required contrast enhancement curve or input vs. output characteristics of our pixel model. **(b)** Image after applying the CE pixel model. **(c)** Histogram of the image of (b).

We want to observe how the pixel output behaves for different input lighting conditions. From equation (1), we know that at the end of the integration time, the amount of voltage drop at PD node ($V_{DD}-V_{PD}$) is proportional to the generated photocurrent of the photodiode. The photocurrent is proportional to the illumination level of the incident light. So, varying the voltage drop at PD node ($V_{DD}-V_{PD}$) from 0 to $V_{DD}$ (=1.2 V) will mimic all the input lighting conditions (from low-level to high-level illumination). The voltage at $V_{OUT}$ is quantized into 256 levels (0-255) for 8-bit encoding. To represent these levels, we vary the voltage drop at PD node, observe the output and normalize it between 0-255 levels, which is shown in Fig. 3(b). We exploit this input vs output characteristics to make a look-up table-based model of this single IPFE pixel and use this model to observe the effect on all the pixels of an array. In this model, we also normalize and quantize the voltage drop at PD node between 0-255 levels so that we can treat this as input intensity. Although the curve in Fig. 3(b) is the input vs output characteristics of our designed circuit, we will call this curve as contrast enhancement or transformation curve (aligns with digital image processing terms) in the subsequent sections for ease of understanding.

Now, we want to see the contrast enhancement performance of the whole pixel array. We take a low-contrast grayscale image (Fig. 3(c))[16], which we will treat as an image taken from a conventional image sensor chip. Then we take each pixel value of this low-contrast image as an input of our pixel model and find the output. The internal meaning of this approach is- a specific pixel value of a conventional image sensor corresponds to a certain level of input illumination. We are connecting the output level of our pixel model to that specific pixel value of the conventional image sensor; thus we are indirectly connecting our model's output level to the input illumination level. This is done to get an idea of how our image sensor would perceive the same scene. After mapping according to our model, the image we get is shown in Fig. 3(d). The low-contrast image captured by the conventional image sensor (Fig. 3(c)) shows that the objects blend indistinguishably with the background. In contrast, the image captured by the IPFE pixel sensors reveals the objects clearly, standing out from the background. This improvement is further supported by the histograms for both images, as shown in Fig. 3(e) and (f). The histogram covers a very narrow range for the low-contrast image; on the other hand, the range expands in the case of the IPFE pixel array. We can calculate the Michelson Contrast Ratio (CR) using the following equation[35]-

$$CR = \frac{L_{max} - L_{min}}{L_{max} + L_{min}} \qquad (2)$$

Where, $L_{max}$ and $L_{min}$ represent the maximum and minimum grayscale level of an image, respectively. For the image without IPFE, the values of $L_{min}$ and $L_{max}$ are 131 and 176, respectively, which results in a CR of 0.15. On the other hand, $L_{min}$ and $L_{max}$ values are 14 and 255, respectively, for the image with IPFE, which leads to a CR of 0.896. This is almost a 6X improvement in CR. However, the CR may not improve if the threshold intensity level does not fall between $L_{min}$ and $L_{max}$ of the input imaging conditions. However, this is not a matter to worry about because the threshold intensity level can be customized according to requirements, which is discussed in the next section.

**Customizing the curve: Contrast Enhancement (CE)**

The ability to customize the contrast enhancement or transformation function of Fig. 1(b) according to specific applications is very crucial. If the image sensor chip is designed for a specific lighting condition, it may not work properly in a very different lighting condition. Moving the points A and B along the x and y directions of the graph provides the ability to customize the function. To start analyzing the customizability of our designed circuit, we set the hypothetical PTM simulation parameters ($R_{HRS}$= 80 kΩ, $R_{LRS}$= 40 kΩ, $I_{C\text{-}HLT}$=4 μA, and $I_{C\text{-}LHT}$=6.8 μA) in such a way that the transformation curve of our designed circuit matches best with a typical contrast enhancement function (such as Fig. 1(b)). The simulated transformation curve of such circuit is shown in

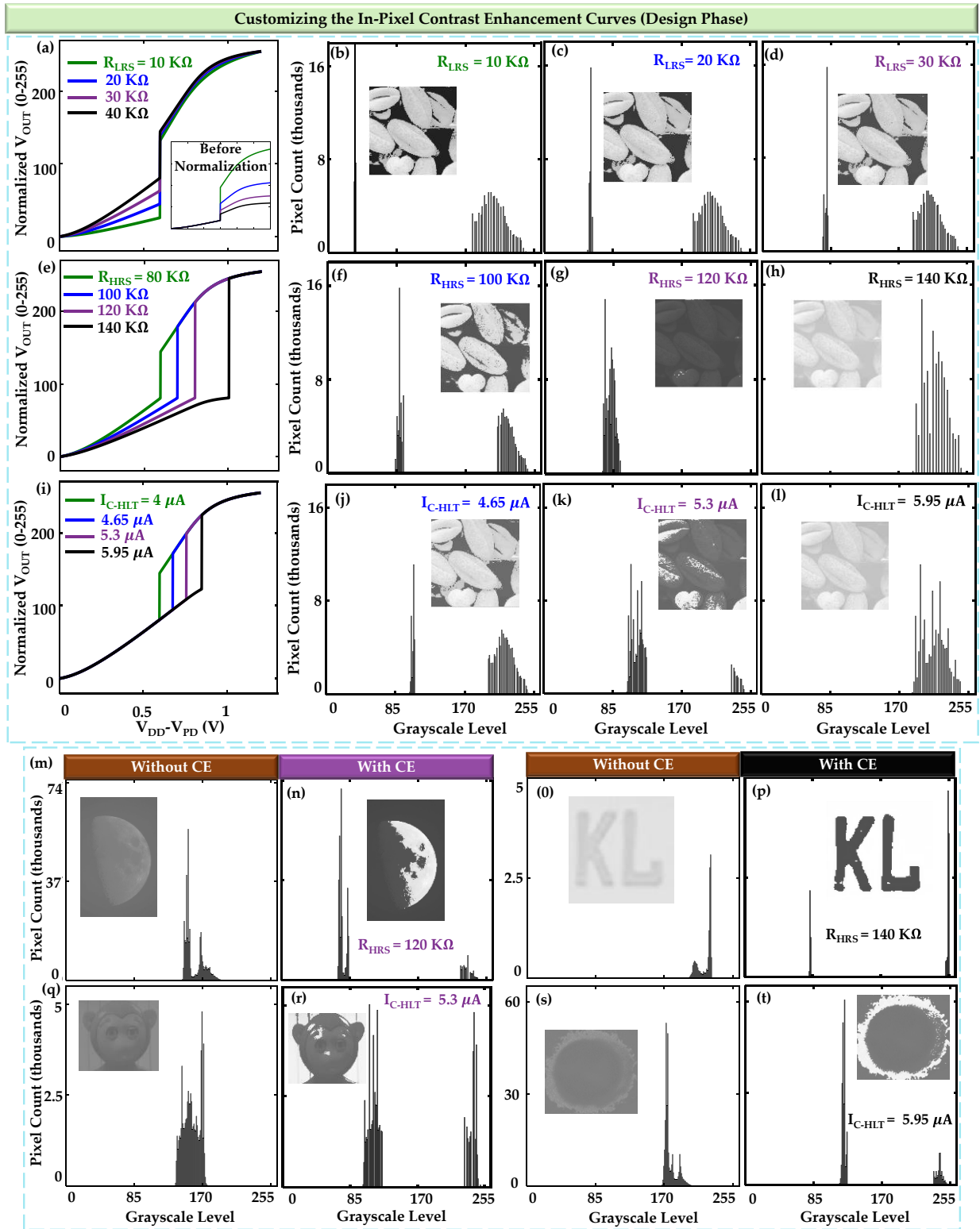

**Figure 5.** Design-phase customization of contrast enhancement mode. (a) The input vs output curves of the pixel model for different resistance values of LRS. The inset shows the curves before normalization. (b)-(d) Histogram and images (inset) after applying the varied pixel models. (e) The curves for different resistance values of HRS (The curves before normalization have not been shown because the maximum and minimum values are the same for all curves, even prior to normalization). (f)-(h) Histogram and images (inset) after applying the varied pixel models. (i) The curves for different $I_{C-HLT}$ values. (j)-(l) Histogram and images (inset) after applying the varied pixel models. The effect of the CE pixel models on other images is illustrated in (m)–(t). (m), (q), (o), and (s) display the histograms without contrast enhancement (CE) pixels, with the insets showing the corresponding images. (n), (r), (p), and (t) present the histograms with CE pixels, and the insets show the effect on the corresponding images.

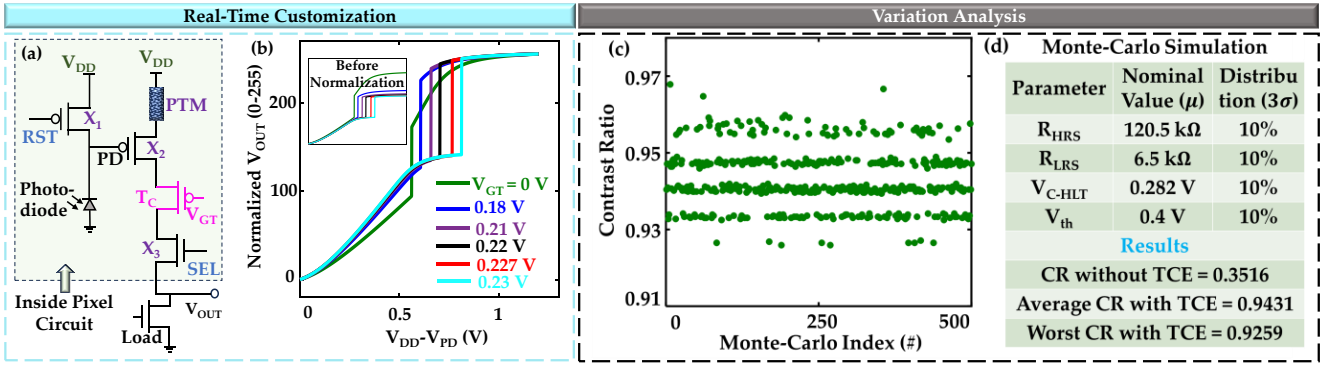

**Figure 6.** Real-time customization of contrast enhancement mode and variation analysis. **(a)** The modified pixel circuit for real-time tuning. **(b)** The input vs output curves of the pixel model for different gate voltages of the added transistor. The inset shows the curves before normalization. **(c)** Michelson contrast ratios for 500 pixel models are calculated and plotted against the Monte-Carlo index. **(d)** Monte-Carlo simulation parameter information and the result summary.

Fig. 4(a). We create another pixel model according to this curve. To see the effect of this transformation curve we apply the pixel model to each pixel of the same low-contrast image of Fig. 3(c) in a similar way discussed in the previous section. The resultant image and the corresponding histogram are shown in Fig. 4 (b) and (c) respectively. Now, we want to move the points A and B of Fig. 4 (a) along the x and y directions (i.e., changing the values of Ax, Ay, Bx, and By) of the graph to reconfigure the transformation function according to arbitrary requirements.

**Design phase customization.** The location of these points in the graph is largely dependent on the PTM parameters. $R_{HRS}$ and $R_{LRS}$ control the slope of stage 1 and stage 2, respectively. The point at which the high to low transition in resistance will occur is determined by $I_{C-HLT}$. It is to be noted that the other critical current, $I_{C-LHT}$ does not have a direct impact on the transformation curve; rather, it only comes into action during the reset phase. One limitation of our designed circuit is that we cannot change Ax and Bx separately because the change in resistance is fully abrupt (i.e. Ax=Bx). By maintaining this condition, the transformation curve can move in any direction, which is shown in Fig. 5.

To change Ay, we vary the $R_{LRS}$ by keeping other parameters fixed. We update our pixel circuit model with each $R_{LRS}$ (Fig. 5(a)) and set the same low-contrast image of Fig. 3(c) as input. We observe the effect of our designed circuit on imaging in a similar fashion described in the previous section. The results are shown in Fig. 5 (b)-(d). To change Ax, Bx, and By, we vary the value of $R_{HRS}$ (Fig. 5(e)) and observe the effect (Fig. 5(f)-(h)). And finally, if we change $I_{C-HLT}$ (Fig. 5(e)), both points A and B move in both x and y directions (Ax, Ay, Bx, By all vary). The observed effects on the image sensor are shown in Fig. 5(j)-(l). The effect on the histogram and image for the combination $R_{LRS}$=40 kΩ, $R_{HRS}$=80 kΩ and $I_{C-HLT}$ =4µA is not shown in Fig. 5 because it is already shown in Fig. 4.

We can observe from the images of Fig. (b)-(l) that not all contrast enhancement curves are suitable for the specific image of Fig. 3(c), which is obvious because the threshold points of all the curves do not align appropriately with the image's requirements. Different contrast enhancement curves will be well-suited for different imaging conditions. That's why we used images captured under different imaging conditions to observe the impact of our designed circuit (Fig. 5(m)-(t)).

It is feasible to obtain a PTM with the required parameters to generate a specific contrast enhancement curve. That said, it is feasible in most cases because of two reasons. Firstly, There are many experimentally demonstrated PTMs with various sets of materials that have different parameters [36–41]. Secondly, if those are not enough for a specific imaging condition, we can modify the dimension of a PTM to meet the specific requirements of the device parameters.

**Real-time customization.** The contrast enhancement curves can also be tuned dynamically during the circuit operation. To achieve that, we modify our pixel circuit (Fig. 6(a)). We incorporate a PMOS transistor, referred to as $T_c$, between the HyperFET and the selector transistor. By varying the gate signal $V_{GT}$ of $T_c$, the contrast enhancement curve can be reconfigured (Fig. 6(b)). According to a specific imaging condition, we can set a certain amount of $V_{GT}$ to achieve a specific contrast enhancement curve.

**Variation analysis**

To observe the effect of variation in the HyperFET of our designed circuit, we perform a 500-point Monte-Carlo (MC) simulation in HSPICE and generate 500 input vs. output curves. We made 500 models of the pixel circuit from those curves and observed the effect on histograms for each of these models in a similar fashion described in the previous sections. We calculate the Michelson Contrast Ratio (CR) using equation (2) for each of the models for a specific image and plot against the MC index (Fig. 6(c)). The parameters of the Monte Carlo (MC) simulation and a summary of the results are shown in Fig. 6(d). The nominal threshold voltage is chosen based on [42,43] and the PTM parameters variation range is chosen from [34].

**Outlook and discussion**
The paper introduces a novel in-pixel contrast enhancement circuit that performs real-time image processing within the pixel array itself, significantly reducing the burden on external processing units, which is crucial for edge devices. Notably, this shift does not increase the complexity of the pixel circuit, as only one additional device is integrated on top of a transistor's source, maintaining efficiency while achieving parallel pixel-level processing. This in-pixel design boosts image quality with a sixfold improvement in Michelson Contrast Ratio (CR) in foreground enhancement mode. While this work focuses on contrast enhancement, it lays the foundation for future developments in fully secure pixel circuits. The enhanced contrast provided by this design simplifies decision-making processes, and the next major advancement could be integrating decision-making capabilities within the pixel circuit itself. The potential for decision-making pixels is unlocked by the high-contrast images produced by this innovative design, making it a key stepping stone toward smarter and more secure imaging solutions.

**Data availability**
The data that support the plots within this paper and other findings of this study are available from the corresponding author upon reasonable request.